\documentclass[onecolumn]{robbyant}           

\usepackage{longtable} 

\metadata[Website]{\url{https://technology.robbyant.com/lingbot-vla}}
\metadata[Github]{\url{https://github.com/robbyant/lingbot-vla}}
\metadata[Checkpoints]{\url{https://huggingface.co/collections/robbyant/lingbot-vla}}
\newcommand{\methodname}{LingBot-VLA}
\newcommand{\method}{\texttt{\methodname}\xspace}

\title{A Pragmatic VLA Foundation Model}

\author{
\begin{center}
    Wei Wu$^{*}$,
    Fan Lu$^{*}$,
    Yunnan Wang$^{*}$,
    Shuai Yang$^{*}$,
    Shi Liu$^{*}$,
    Fangjing Wang$^{*}$,
    Qian Zhu,
    He Sun,
    Yong Wang,
    \\[3pt]
    Shuailei Ma,
    Yiyu Ren,
    Kejia Zhang,
    Hui Yu,
    Jingmei Zhao,    
    Shuai Zhou,
    Zhenqi Qiu,
    Houlong Xiong,
    \\[3pt]
    Ziyu Wang,
    Zechen Wang,
    Ran Cheng,
    Yong-Lu Li,
    Yongtao Huang,
    Xing Zhu,
    Yujun Shen,
    Kecheng Zheng$^{\dagger}$
    \\[12pt]
    {$^{*}$Equal Contribution} \qquad
    {$^{\dagger}$Project Lead}
\end{center}
}

\begin{document}

\vspace{-30pt}

\abstract{%

%
Offering great potential in robotic manipulation, a capable Vision-Language-Action (VLA) foundation model is expected to faithfully generalize across tasks and platforms while ensuring cost efficiency (\textit{e.g.}, data and GPU hours required for adaptation).
To this end, we develop \method with around 20,000 hours of real-world data from 9 popular dual-arm robot configurations.
Through a systematic assessment on 4 robotic platforms, each completing 100 tasks with 130 post-training episodes per task, our model achieves clear superiority over competitors, showcasing its \textbf{\textit{strong performance}} and \textbf{\textit{broad generalizability}}.
We have also built an \textbf{\textit{efficient}} codebase, which delivers a throughput of $261$ samples per second with an 8-GPU training setup, representing a $1.5\sim2.8\times$ (depending on the relied VLM base model) speedup over existing VLA-oriented codebases.
The above features ensure that our model is well-suited for real-world deployment.
To advance the field of robot learning, we provide open access to the code, base model, and benchmark data, with a focus on enabling more challenging tasks and promoting sound evaluation standards.
%
%
%
%
%
%

}

\maketitle

\justifying
\begin{figure}[t]
\centering
\includegraphics[width=\linewidth]{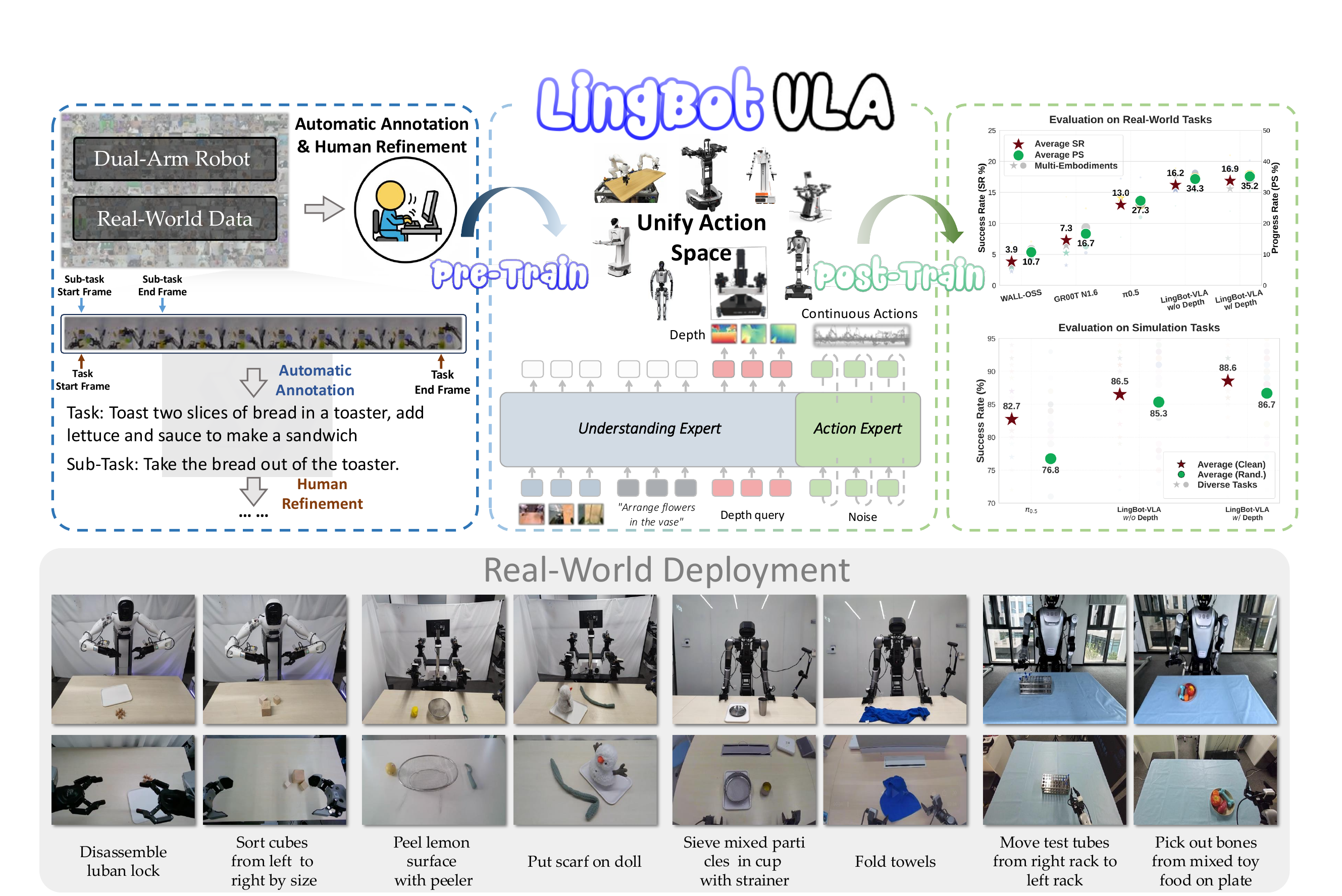}
\caption{\textbf{Overview} of \method. We scale dual-arm robot data collected in the real world for pre-training. \method can be easily and efficiently transferred to downstream tasks. Moreover, we conduct a systematic assessment across three robotic embodiments, which demonstrates the clear superiority of our model.}
\label{fig:teaser}
\end{figure}

\section{Introduction}\label{sec:intro}

Vision-Language-Action (VLA) foundation models~\cite{pi_0,pi_0d5,gr00t_n1_6} have emerged as a promising method for enabling robots to perform diverse manipulation tasks guided by natural language instructions. 
Through large-scale pre-training, these models acquire generalizable skills that can be rapidly adapted to diverse tasks and robotic platforms.
Despite the significant progress, there remains a lack of comprehensive empirical studies on how real-robot performance scales with increasingly vast pre-training datasets. Moreover, the community lacks a highly optimized training codebase capable of efficiently conducting these scaling evaluations on massive volumes of data.
%
Consequently, a fundamental question that demands investigation in the real-world setting is: \textit{How do VLA models truly scale with massive real-world robot data?}

Understanding the scaling behavior of VLA models is crucial for robotic learning, especially on vast and diverse real-world datasets.
%
%
In this work, we provide a systematic empirical investigation into how success rates scale with respect to data volume and diversity during VLA pre-training
By scaling pre-training data from 3,000 hours to 20,000 hours, we demonstrate that downstream success rates improve consistently and substantially. Notably, this scaling behavior shows no signs of saturation even at the 20,000-hour mark, suggesting that VLA performance continues to benefit from increased data volume. These results provide the first empirical evidence of favorable scaling properties in real-world robot learning, offering critical insights for future VLA development and large-scale data curation.

While scaling analysis reveals favorable performance trends, translating these insights into reliable, deployable systems necessitates rigorous evaluation on real robotic platforms at a large scale. Thanks to GM-100~\cite{gm100}, which provides 100 carefully designed tasks, we conduct a systematic assessment across 4 robotic platforms, involving 130 episodes per task per embodiment. 
By emphasizing task diversity and multi-platform consistency, our evaluation framework provides a choice of new standards for sound VLA benchmarking.
%


In this paper, we present \method, a pragmatic VLA foundation model trained on about 20,000 hours of real-world manipulation data from 9 robotic platforms. 
Our systematic evaluation on the comprehensive benchmark, demonstrates that \method achieves state-of-the-art performance and exceptional generalization compared to existing methods. Beyond model capabilities, we emphasize that large-scale robot learning necessitates high computational efficiency. To this end, we have developed an optimized codebase that achieves a throughput of 261 samples per second on an 8-GPU cluster. This efficiency gain substantially shortens training cycles and reduces computational overhead, thereby lowering the overall costs. By combining superior performance, broad generalizability, and computational efficiency, \method is well-positioned for real-world robotic applications. To foster community progress, we provide open access to the code, base model, and benchmark data, with a focus on enabling more challenging tasks and promoting sound evaluation standards.
%

\section{Related Work}\label{sec:related}

\subsection{Vision-Language-Action Models}
\textbf{Foundation VLA.} 
Vision-language-action foundation models typically adopt a powerful pre-trained vision-language model~\cite{bai2025qwen2, paligemma} as the semantic backbone, coupled with diffusion-based action head.
Recent VLA foundation models~\cite{pi_0, pi_0d5, gr00tN1, gr00t_n1_6, gemini_robotics, walloss, galaxeaG0, magma, gr3} have demonstrated enhanced multi-task execution capabilities and superior multi-embodiment adaptability, following pre-training on larger-scale and increasingly diverse datasets.
Distinguishing itself from the datasets utilized in preceding VLA foundation models, our model is pre-trained on an extensive corpus approximate 20,000 hours of multi-embodiment data. This massive-scale dataset, characterized by its high behavioral diversity, significantly bolsters the model's generalization capabilities across various robotic manipulation tasks.

\noindent
\textbf{Spatial VLA.} 
While traditional VLA models excel at semantic understanding, they often struggle with precise geometric reasoning and depth perception required for complex spatial manipulation.
To address this, several works~\cite{spatialvla, Spatial_forcing, internvla_M1, guo2025omnivla, magma, gemini_robotics_1d5, geovla} have integrated spatial representations into the VLA framework.
Several research~\cite{magma, internvla_M1, gemini_robotics} initiatives have focused on bolstering the spatial awareness of VLMs within embodied scenarios to enhance the spatial manipulation capabilities of VLAs in downstream tasks. 
Others explicitly or implicitly incorporate depth information during the VLA training phase.
Spatial Forcing~\cite{Spatial_forcing} employs a streamlined alignment strategy that compels the integration of VLA visual embeddings with spatial representations, thereby significantly improving the model's spatial comprehension.

\subsection{Evaluation on Robot Policy}
Current evaluation methodologies for robot policies are primarily bifurcated into two categories: simulation-based~\cite{libero, calvin, simplerenv, robocasa, robotwin2} and real-world embodiment-based~\cite{robochallenge, roboarena}.
Simulation-based benchmark provide a rapid and convenient means to evaluate the capabilities of policies, enabling large-scale parallel testing across vast and diverse interaction scenarios at very low cost. 
Although simulation environments typically employ idealized physical models, their results often do not fully represent the complexity of the real physical world.
The another real-world evaluations' efficiency is often bottlenecked by the requirement for extensive hardware parallelism. 
Consequently, the majority of prior VLA studies have been confined to comparing a limited number of methods across only a few tasks. 
To more comprehensively evaluate the real-world performance of policies, this work conducts an assessment across three distinct robotic platforms, with 100 tasks executed on each platform. We further provide a thorough analysis of how mainstream VLA models adapt to the diversity encountered in real-world scenarios.

\subsection{Efficient VLA Training}
The rapid iteration of VLA models has catalyzed the development of specialized training infrastructure. Several well-designed open-source codebases have recently emerged in the community, each catering to different research priorities. 
For instance, the OpenPI~\cite{pi_0} repository provides a versatile framework supporting both JAX and PyTorch for training the $\pi$ series models. 
StarVLA~\cite{starvla} introduces a modular and user-friendly codebase specifically optimized for the co-training of VLAs and VLMs, facilitating the transfer of semantic knowledge to robotic control. 
Additionally, Dexbotic~\cite{dexbotic} is designed as a unified and efficient solution to streamline the development lifecycle of VLAs, focusing on standardizing the pipeline from data ingestion to model deployment.
Despite these advancements, training large-scale VLA models on multi-node clusters remains a significant challenge due to data I/O bottlenecks and communication overheads.
To bridge this gap, we present \method, a high-performance open-source codebase engineered for large-scale VLA training. Unlike existing frameworks, our codebase implements systemic optimizations in data loading, distributed training strategies, and operator-level acceleration. 
These enhancements lead to a comprehensive improvement in training throughput and scalability, providing a more efficient foundation for the community to explore the scaling limits of robotic foundation models.


\section{Pre-training Dataset}\label{sec:method}

\begin{figure}[t]
    \centering
    \includegraphics[width=\linewidth]{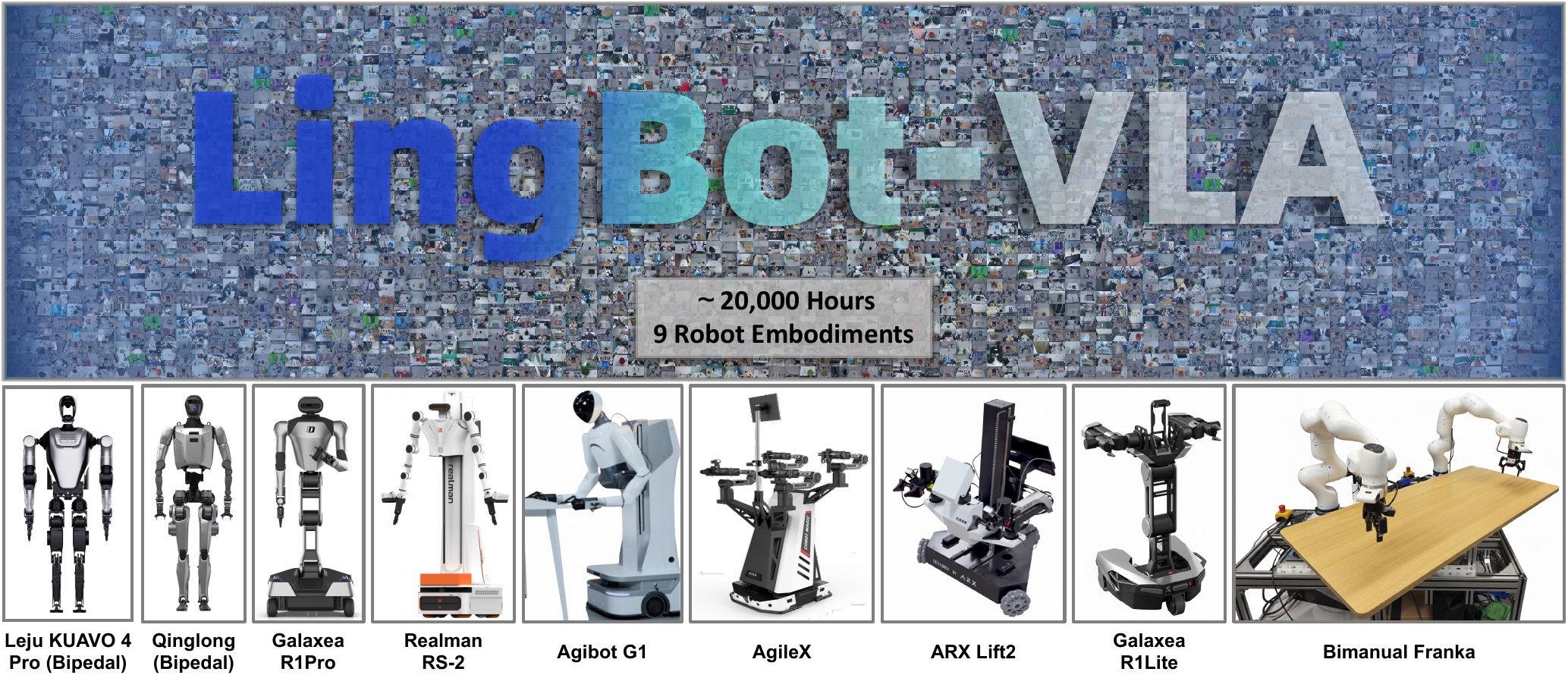}
    \caption{\textbf{Visualization of pre-training dataset} used by \method.}
    \label{fig:data}
\end{figure}

\subsection{Data Collection}
The pre-training dataset is built upon large-scale teleoperated data collected from 9 popular dual-arm robot embodiments, as shown in~\cref{fig:data}. We discuss these embodiments below:

\begin{itemize}

    \item\textbf{AgiBot G1.}
    This setup has two 7-DoF arms with three RGB-D cameras. Robot data are collected via VR-based teleoperation on this setup.
    
    \item\textbf{AgileX.}
    This setup is equipped with three cameras and two 6-DoF arms. Robot control is achieved using isomorphic arms during the data collection process.

    \item\textbf{Galaxea R1Lite.}
    This setup has two 6-DoF arms, with one stereo camera and two wrist cameras.
    
    \item\textbf{Galaxea R1Pro.}
    Two 7-DoF arms, one stereo camera, and two wrist cameras are used in this setup. 
    
    \item\textbf{Realman Rs-02.}
    This setup uses three cameras and features a 16-dimensional configuration and action space: two 7-DoF arms and two parallel grippers.
    
    \item\textbf{Leju KUAVO 4 Pro.}
    A bipedal humanoid robot featuring two 7-DoF arms, two parallel grippers, one head-mounted camera, and two wrist-mounted cameras.
    
    \item\textbf{Qinglong.}
    A humanoid robot with two 7-DoF arms and three cameras: one on the head and one on each wrist.
    
    \item\textbf{ARX Lift2.}
    This setup uses three cameras and two 6-DoF arms.
    
    \item\textbf{Bimanual Franka.}
    This setup uses two 7-DoF arms and two parallel grippers, forming a 16-dimensional action space, with three cameras.
     
\end{itemize}










\subsection{Data Labeling}
To obtain precise language instructions, we perform the following annotations: 
(1) \textit{Video Segment}.
Videos from multiple viewpoints, captured by robots, are jointly decomposed into clips by human annotators according to predefined atomic actions. 
Besides, to reduce the redundant information within videos, static frames at the start and end of the videos are eliminated at this stage.
(2) \textit{Instruction Annotation}. 
After obtaining videos containing the robots' full motion trajectories and video clips for each atomic action, we employ Qwen3-VL-235B-A22B~\cite{bai2025qwen2} for precise annotation of task and sub-task instructions, as shown in~\cref{fig:teaser}.

\section{Model Training}
\subsection{Architecture}
To leverage well-trained vision-language representations, \method integrates the pre-trained VLM (\textit{i.e.}, Qwen2.5-VL~\cite{bai2025qwen2}) with an initialized action generation module called `action expert'. These components are organized via a Mixture-of-Transformers (MoT) architecture like BAGEL~\cite{deng2025emerging}, where vision-language and action modalities are processed through distinct transformer pathways, coupled by a shared self-attention mechanism for layer-wise unified sequence modeling. This MoT framework ensures that high-dimensional semantic priors from the VLM provide continuous guidance across all layers, while simultaneously mitigating cross-modal interference by maintaining modality-specific processing. The architecture of \method is illustrated in~\cref{fig:teaser}. Multi-view operational images and the related task instruction are uniformly encoded through a VLM to establish multimodal conditioning for subsequent action generation. Concurrently, the robot's proprioceptive sequences, specifically initial states and action chunks, are fed into the action expert for the prediction of action generation. We employ Flow Matching~\cite{lipman2022flow} for continuous action modeling, which facilitates fluid and smooth robotic control, ensuring high-precision execution across complex tasks and diverse robots.

In \method, the VLM and the action expert interact through a shared self-attention mechanism, facilitating a unified layer-wise representation. Consequently, the joint modeling sequence at timestamp $t$ is formulated as the concatenation of the observation conditions $\mathbf{O}_t$ and the action chunk $\mathbf{A}_t$. 
Specifically, the observation context is defined as:
\begin{equation}
    \mathbf{O}_t = [\mathbf{I}_t^1, \mathbf{I}_t^2, \mathbf{I}_t^3, \mathbf{T}_t, \mathbf{s}_t],
\end{equation}
which incorporates tokens from three-view operational images $\mathbf{I}_t^{1,2,3}$ of dual-arm robots, the task instruction $\mathbf{T}_t$, and the robot state $\mathbf{s}_t$. The corresponding action sequence is denoted as:
\begin{equation}
    \mathbf{A}_t = [\mathbf{a}_t, \mathbf{a}_{t+1}, \dots, \mathbf{a}_{t+T-1}],
\end{equation}
where $T$ represents the action chunk length, \textit{i.e.}, the temporal horizon of the predicted trajectory, which is set to 50 during our pre-training stage. Therefore, the training objective is to characterize the conditional distribution $p(\mathbf{A}_t | \mathbf{O}_t)$ through conditional flow matching. For a flow timestep $s \in [0, 1]$, we define a probability path through linear interpolation between the Gaussian noise $\epsilon \sim \mathcal{N}(\mathbf{0}, \mathbf{I})$ and the ground-truth action $\mathbf{A}_t$, obtaining the intermediate action $\mathbf{A}_{t, s}=s\mathbf{A}_{t}+(1-s)\epsilon$. The conditional distribution of $\mathbf{A}_{t, s}$ is formulated as:
\begin{equation}
    p(\mathbf{A}_{t, s} | \mathbf{A}_t) = \mathcal{N}(s\mathbf{A}_{t}, (1-s)\mathbf{I}).
\end{equation}
The action expert $v_\theta$ is trained to predict the conditional vector field by minimizing the Flow Matching objective:
\begin{equation}
    \mathcal{L}_{\text{FM}} = \mathbb{E}_{s \sim \mathcal{U}[0, 1], \mathbf{A}_t, \mathbf\epsilon} \left\| v_\theta(\mathbf{A}_{t, s}, \mathbf{O}_t, s) - (\mathbf{A}_t - \mathbf\epsilon) \right\|^2,
\end{equation}
where the target velocity is given by the ideal vector field $\mathbf{A}_t - \mathbf\epsilon$ derived from the linear probability path.

Following $\pi_0$~\cite{pi_0}, we implement blockwise causal attention for modeling the joint sequence $[\mathbf{O}_t, \mathbf{A}_t]$. The sequence can be partitioned into three distinct functional blocks: $[\mathbf{I}_t^1, \mathbf{I}_t^2, \mathbf{I}_t^3, \mathbf{T}_t]$, $[\mathbf{s}_t]$ and $[\mathbf{a}_t, \mathbf{a}_{t+1}, \dots, \mathbf{a}_{t+T-1}]$.
A causal mask is applied among these blocks, such that tokens in each block can only attend to themselves and those in preceding blocks. Conversely, all tokens within the same block employ bidirectional attention and can attend to each other. This configuration ensures that the action expert can leverage all available observation knowledge, while preventing information leakage from future action tokens into the current observation representations.

To explicitly capture spatial awareness within manipulation environments and further enhance the robot's execution robustness, we adopt a vision distillation approach inspired by recent works~\cite{wang2025vision,huang2025mllms}. Specifically, we apply the learnable queries $[\mathbf{Q}^1_t, \mathbf{Q}^2_t,\mathbf{Q}^3_t]$ corresponding to three-view operational images. To integrate depth information, these queries are processed by VLM and then aligned with the depth tokens $[\mathbf{D}^1_t, \mathbf{D}^2_t,\mathbf{D}^3_t]$ from LingBot-Depth~\cite{lingbotdepth}. 
We align the VLM learnable queries and LingBot-Depth tokens by minimizing the distillation loss $\mathcal{L}_{distill}$:
\begin{align}
\mathcal{L}_{distill} =\mathbb{E}_{\mathbf{Q}_t}\left|{ \text{Proj}(\mathbf{Q}_t)} - \mathbf{D}_t\right|,
\label{eq::2}
\end{align}%
where $\text{Proj}(\cdot)$ is a projection layer that applies cross-attention for dimensional alignment. This integration infuses geometric information into the \method model, enabling precise perception for complex manipulation tasks.

\subsection{Training Efficiency Optimization}

Given that action data is inherently high-frequency, establishing a highly efficient pipeline encompassing distributed training and operator optimization is imperative. Our optimization methodology is structured as follows:


\textbf{Distributed Strategy:} While VLA models typically possess a moderate parameter count, achieving an optimal trade-off between GPU memory occupancy and training throughput remains essential. We employ Fully Sharded Data Parallel (FSDP)—a highly efficient PyTorch implementation of the Zero Redundancy Optimizer (ZeRO)—to shard optimizer states, model parameters, and gradients, thereby minimizing memory footprint. Drawing inspiration from the Hybrid Sharded Data Parallel (HSDP) approach proposed in VeOmni~\cite{ma2025veomni}, we construct specific ``shard groups'' exclusively for the action expert modules. This strategy effectively mitigates the communication overhead associated with excessive parameter sharding. Additionally, we implement a mixed-precision policy: performing reductions in \texttt{torch.float32} to ensure numerical stability, while utilizing \texttt{torch.bfloat16} for storage and communication.

\textbf{Operator-Level Optimization:} The multimodal fusion of vision, language, and action within our architecture is fundamentally a sparse attention process. To address this, we leverage FlexAttention to optimize computation. Furthermore, we apply operator fusion (via \texttt{torch.compile}) to reduce kernel launch overhead and maximize memory bandwidth utilization.

\section{Experiments}\label{sec:exp}

\subsection{Large-scale Real-world Benchmark}
We conduct a large-scale empirical evaluation of \method designed to rigorously assess multi-embodiment generalization and real-world robustness. Our experimental framework comprises three core components: (1) 25 physical robots spanning 4 distinct commercial platforms, (2) GM-100~\cite{gm100} benchmark featuring 100 diverse manipulation tasks with 39,000 expert demonstrations 
and (3) a controlled evaluation protocol generating 22,500 trials 
comparing \method against three state-of-the-art baselines under identical training and testing conditions.

\subsubsection{Hardware Platforms}
We conduct experiments across 4 distinct robotic platforms: AgileX, Agibot G1 Galaxea R1Pro and Leju KUAVO 4 Pro. 
All three embodiments feature a dual-arm configuration equipped with parallel-jaw grippers.
To ensure robust perception, each robot is outfitted with multiple cameras: two wrist-mounted cameras and a head-mounted camera to capture an egocentric, human-eye perspective. 
All tasks are tabletop-based, with the embodiment's chassis and waist securely fixed in place.


\subsubsection{Data Collection and Processing}

\begin{figure}[t]
    \centering
    \begin{minipage}{0.48\linewidth}
        \centering
        \includegraphics[width=\linewidth]{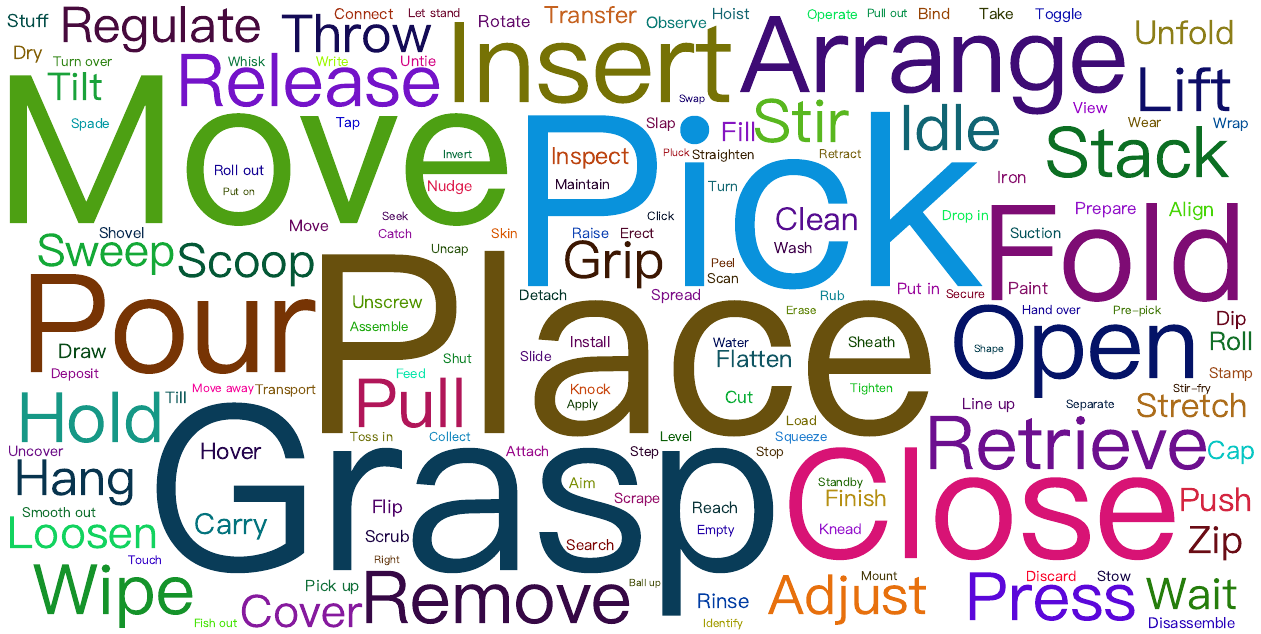}
        \subcaption{}
        \label{fig:train_data_action}
    \end{minipage}
    \hfill
    \begin{minipage}{0.48\linewidth}
        \centering
        \includegraphics[width=\linewidth]{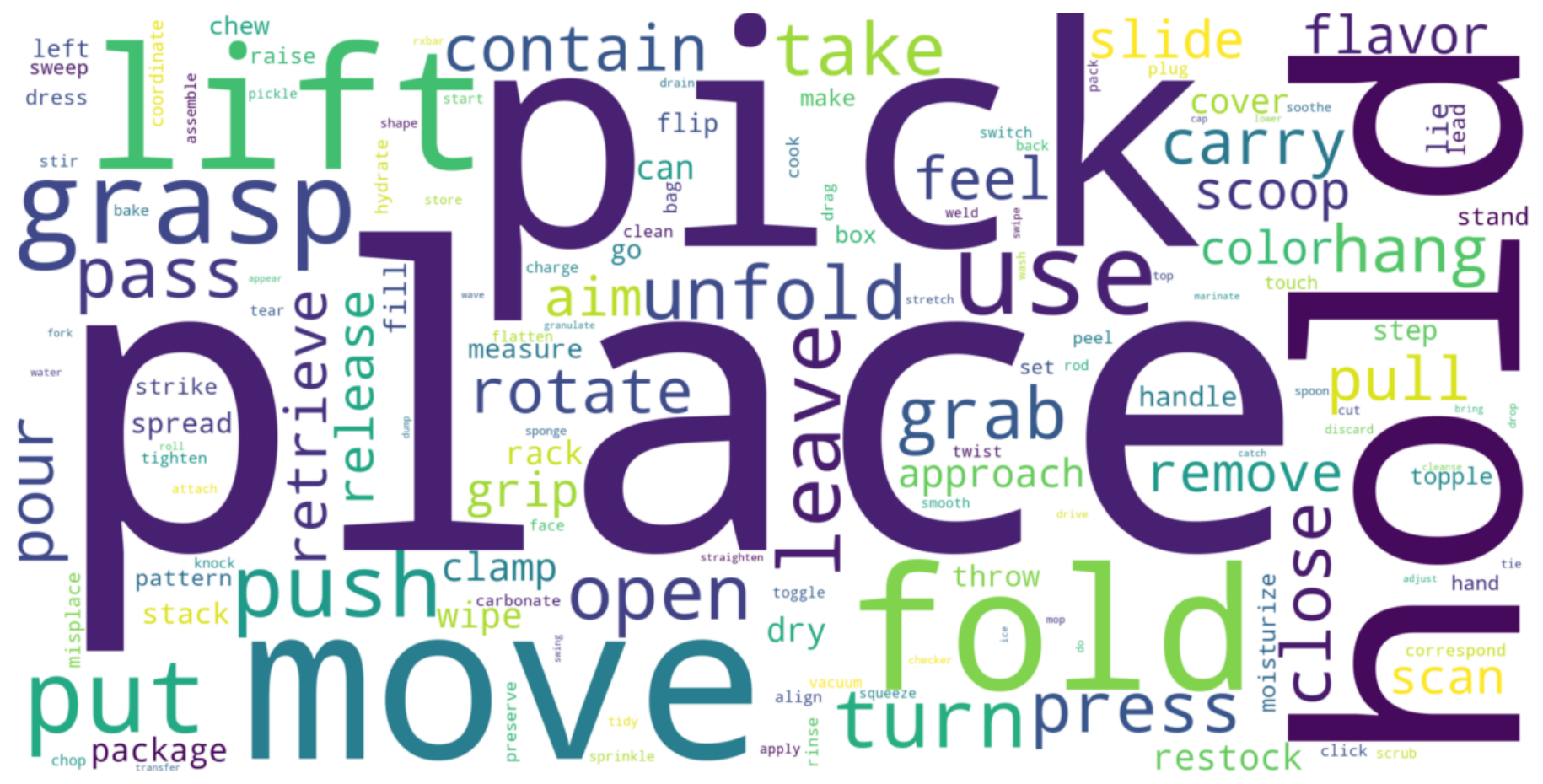}
        \subcaption{}
        \label{fig:benchmark_data_action}
    \end{minipage}
    \caption{
    \textbf{Word cloud of atomic actions} in (a) Pre-training datasets and (b) Benchmark.}
\end{figure}


For each GM-100 task~\cite{gm100}, we collect expert demonstrations via teleoperation following a standardized protocol designed to ensure high data quality and environmental diversity.
\textit{Trajectory Volume:}
150 raw trajectories are collected per task across three platforms. The top 130, ranked by execution quality (task completion, motion smoothness, and protocol adherence), are retained for training. All trajectories strictly follow GM-100 task specifications.
\textit{Standardized Objects:}
task objects are standardized and sourced according to GM-100 material specifications to ensure reproducibility across sites.
\textit{Environmental Diversity:}
object poses (positions and orientations) are randomized within the workspace for each trajectory to prevent overfitting to specific spatial configurations and encourage learning of task-relevant invariances.
\textit{Teleoperation Guidelines:}
(1) maintaining clearance between the end-effector and workspace surfaces to avoid collisions,
(2) reducing velocity during object contact phases for smooth manipulation, and
(3) ensuring distinct image observations at episode start and termination for reliable policy training.
\textit{Automated Filtering:} an algorithmic screening procedure automatically excludes episodes exhibiting technical anomalies.
\textit{Manual Review:} 
human reviewers validate the filtered dataset using synchronized multi-view video streams. Episodes are removed if they include extraneous objects or deviate from task protocols.


To analyze the semantic distribution and diversity of action categories, we visualized the most prevalent atomic actions in the training and testing sets using word clouds, as shown in~\cref{fig:train_data_action,fig:benchmark_data_action}. Quantitative analysis reveals that approximately 50\% of the atomic actions in the test set are absent from the top 100 most frequent training actions. This significant discrepancy underscores the diversity of our test set and ensures a rigorous assessment of the model's generalization capabilities.

\subsubsection{Benchmarking and Evaluation Protocol}
We systematically compare \method with three state-of-the-art VLA models: $\pi_{0.5}$, GR00T N1.6, and WALL-OSS, under strict experimental controls to isolate architectural performance.
\textit{Standardized Training:} 
All models are fine-tuned from publicly available pre-trained checkpoints using the same post-training pipeline. The verified dataset (130 filtered trajectories per task) and consistent hyperparameters (\textit{i.e.}, batch sizes=256, epochs=20) are applied to ensure fair comparisons.
\textit{Strict Machine-Task Pairing:} 
To eliminate hardware-induced variance, evaluations are conducted on the exact robot units used during data collection. All models are tested sequentially on the same hardware-task pair in randomized order. For example, in the ``Stack Bowls'' task, all models are evaluated on the same unit across AgileX, Agibot G1, and Galaxea R1pro platforms.
\textit{Controlled Evaluation Setup:} 
Testing conditions follow standardized protocols, mirroring data collection procedures with randomized object positions and orientations while maintaining consistent task specifications. This ensures evaluation of generalization rather than memorization.
\textit{Inference and Recording:} 
Each model undergoes 15 trials per task-robot pair for statistical robustness. Evaluation environments are kept constant, and comprehensive data (\textit{e.g.}, third-person views, robot states, and model predictions) are recorded in rosbag format for transparency. These recordings will be open-sourced to establish verifiable benchmarks.

\subsubsection{Evaluation Metrics}

We evaluate model performance using two metrics capturing both task completion and partial progress.
\textit{Success Rate (SR):} 
the proportion of trials where the model completes all task steps within a 3-minute time limit.
This primary metric reflects the model's real-world deployment viability.
\textit{Progress Score (PS):} 
measures partial task completion by tracking progress through sequential subtask checkpoints:
For example, in a 6-step ``Stack Bowls'' task, completing steps 1–4 but failing at step 5 results in a score of $\frac{4}{6} \approx 0.67$. This diagnostic metric highlights failure modes and rewards partial success.
\textit{Termination Criteria:}
A trial ends if: (1) three consecutive subtask failures occur, or (2) safety-critical events (\textit{e.g.}, collisions) arise. Progress is scored based on subtasks completed before termination.
We report overall SR and PS across 100 tasks, and per-platform metrics stratified by robot type to assess cross-embodiment generalization.

\subsection{Comparison on Real-world Benchmark}~\label{sec:real}
As shown in~\cref{tab:real-world-evaluation}, we compare our two \method variants with three strong baselines across three platforms. On all platforms, \method \textit{w/o} depth significantly outperforms WALL-OSS and GR00T N1.6 in both SR and PS metrics. By incorporating depth-based spatial information, \method\textit{w/} depth achieves an average SR improvement of $4.28\%$ and a PS increase of $7.76\%$ over $\pi_{0.5}$ across the three embodiments.
Notably, GR00T N1.6 performs average on the Agibot G1, AgileX and Leju KUAVO 4 Pro embodiment but achieves SR and PS comparable to $\pi_{0.5}$ on the Galaxea R1Pro platform. This is due to the extensive inclusion of Galaxea R1Pro data during its pre-training, indicating that pre-training can significantly enhance performance on downstream tasks with high structural similarity.
The complete and detailed test results can be found in the Appendix~\cref{tab:exp_AgilexCobotMagic_1}-~\ref{tab:exp_LejuKUAVO_2}.


\begin{table}[t]
  \centering
  \caption{\textbf{Experiment results of real-world evaluation} on GM-100~\cite{gm100} benchmark. `SR' refers to success rate, and `PS' refers to progress score.}
  \vspace{-3pt}
  \resizebox{0.99\linewidth}{!}{
\begin{tabular}{ccccccccccc}
\toprule
              \multicolumn{1}{c}{\multirow{2}[2]{*}{\textbf{Platform}}} & \multicolumn{2}{c}{WALL-OSS} & \multicolumn{2}{c}{GR00T N1.6} & \multicolumn{2}{c}{$\pi_{0.5}$} & \multicolumn{2}{c}{Ours \textit{w/o} depth} & \multicolumn{2}{c}{Ours \textit{w/} depth} \\
              & \textbf{SR}           & \textbf{PS}            & \textbf{SR}          & \textbf{PS}          & \textbf{SR}          & \textbf{PS}         & \textbf{SR}               & \textbf{PS}              & \textbf{SR}          & \textbf{PS}         \\
              \midrule
Agibot G1 (Wheeled)      & 2.99\%       & 8.75\%        & 5.23\%      & 12.63\%     & 7.77\%      & 21.98\%    & \textbf{12.82\%}          & 30.04\%         & 11.98\%     & \textbf{30.47\%}    \\
AgileX  (Wheeled)      & 2.26\%       & 8.16\%        & 3.26\%      & 10.52\%     & 17.20\%     & 34.82\%    & 15.50\%          & 36.31\%         & \textbf{18.93\%}     & \textbf{40.36\%}    \\
Galaxea R1Pro (Wheeled) & 6.89\%       & 14.13\%       & 14.29\%     & 24.83\%     & 14.10\%     & 26.14\%    & 18.89\%          & 34.71\%         & \textbf{20.98\%}     & \textbf{35.40\%}    \\
Leju KUAVO 4 Pro (Bipedal) & 3.26\%       & 11.75\%       & 6.45\%     & 18.66\%     & 12.91\%     & 26.35\%    & \textbf{17.59\%}          &  \textbf{36.22\%}         & 15.60\%     & 34.40\%    \\
\midrule
\textbf{Average} & 3.85\% & 10.70\%  & 7.31\% & 16.66\% & 13.00\% & 27.32\% & 16.20\% & 34.32\% & \textbf{16.87}\% & \textbf{35.16}\%   \\
\bottomrule
\end{tabular}}
\label{tab:real-world-evaluation}
\end{table}


\begin{table}[t]
    \centering
    \caption{\textbf{Experiment results of simulation evaluation} on RoboTwin 2.0~\cite{robotwin2} benchmark.}
    \vspace{-3pt}
    \begin{minipage}{0.49\linewidth}
        \centering
    \subcaption{Clean Scenes}
    \vspace{0.3em}
    \resizebox{\linewidth}{!}{
        \begin{tabular}{cccc}
        \toprule
         & {\textbf{$\pi_{0.5}$}} & {Ours \textit{w/o} depth} & {Ours \textit{w/} depth} \\
        \midrule
        \textbf{Average SR} & 82.74\%  & {86.50\%} & \textbf{88.56\%} \\
        \bottomrule
        \end{tabular}
    }
    \end{minipage}
    \hfill
    \begin{minipage}{0.49\linewidth}
        \centering
        \subcaption{Randomized Scenes}
        \vspace{0.3em}
        \resizebox{\linewidth}{!}{
                \begin{tabular}{cccc}
        \toprule
         & {\textbf{$\pi_{0.5}$}} & {Ours \textit{w/o} depth} & {Ours \textit{w/} depth} \\
        \midrule
        \textbf{Average SR} & 76.76\%  & {85.34\%} & \textbf{86.68\%} \\
        \bottomrule
        \end{tabular}
        }
    \end{minipage}
    \label{tab:robotwin_avg}%
\end{table}




\subsection{Comparison on Simulation Benchmark}~\label{sec:sim}
\noindent In~\cref{tab:robotwin_avg}, we evaluate simulation performance across $50$ representative manipulation tasks within the RoboTwin 2.0 suite. Starting from pretrained checkpoints, each model was further finetuned on the RoboTwin dataset. To assess multi-task generalization, we train all models on $2,500$ demonstrations from clean scenes (50 per task) and $25,000$ from highly randomized scenes (500 per task). Randomization factors encompass varied backgrounds, table-top clutter, table-height perturbations, and diverse lighting conditions. Compared to the $\pi_{0.5}$ baseline, \method demonstrates marked advancements in RoboTwin 2.0 multi-task settings. Specifically, \method \textit{w/o} depth yields absolute success rate increases of over $3.76\%$ in clean environments and $8.58\%$ in randomized scenarios.
By employing learnable query-based alignment, the integration of depth information enables \method to effectively extract rich spatial priors from LingBot-Depth model. The approach surpasses the baseline model by absolute margins of $5.82\%$ and $9.92\%$ in clean and randomized configurations, respectively. Please refer to~\cref{tab:robotwin} of Appendix for detailed results. 

\begin{figure}[t]
    \centering
    \begin{minipage}{0.48\linewidth}
        \centering
        \includegraphics[width=\linewidth]{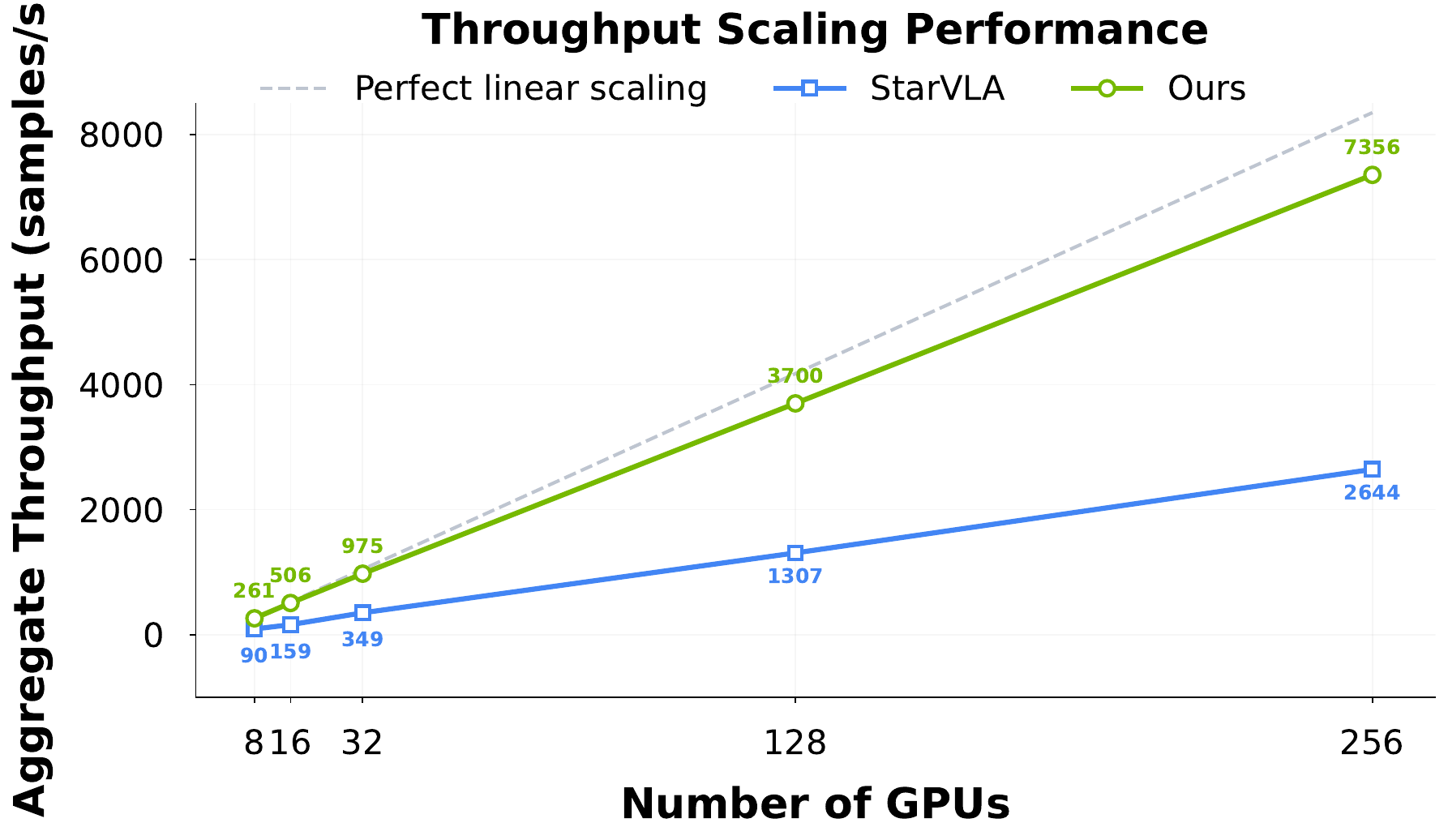}
        \subcaption{Qwen2.5-VL-3B-$\pi$ model}
        \label{fig1:tp_qwenvl}
    \end{minipage}
    \hfill
    \begin{minipage}{0.48\linewidth}
        \centering
        \includegraphics[width=\linewidth]{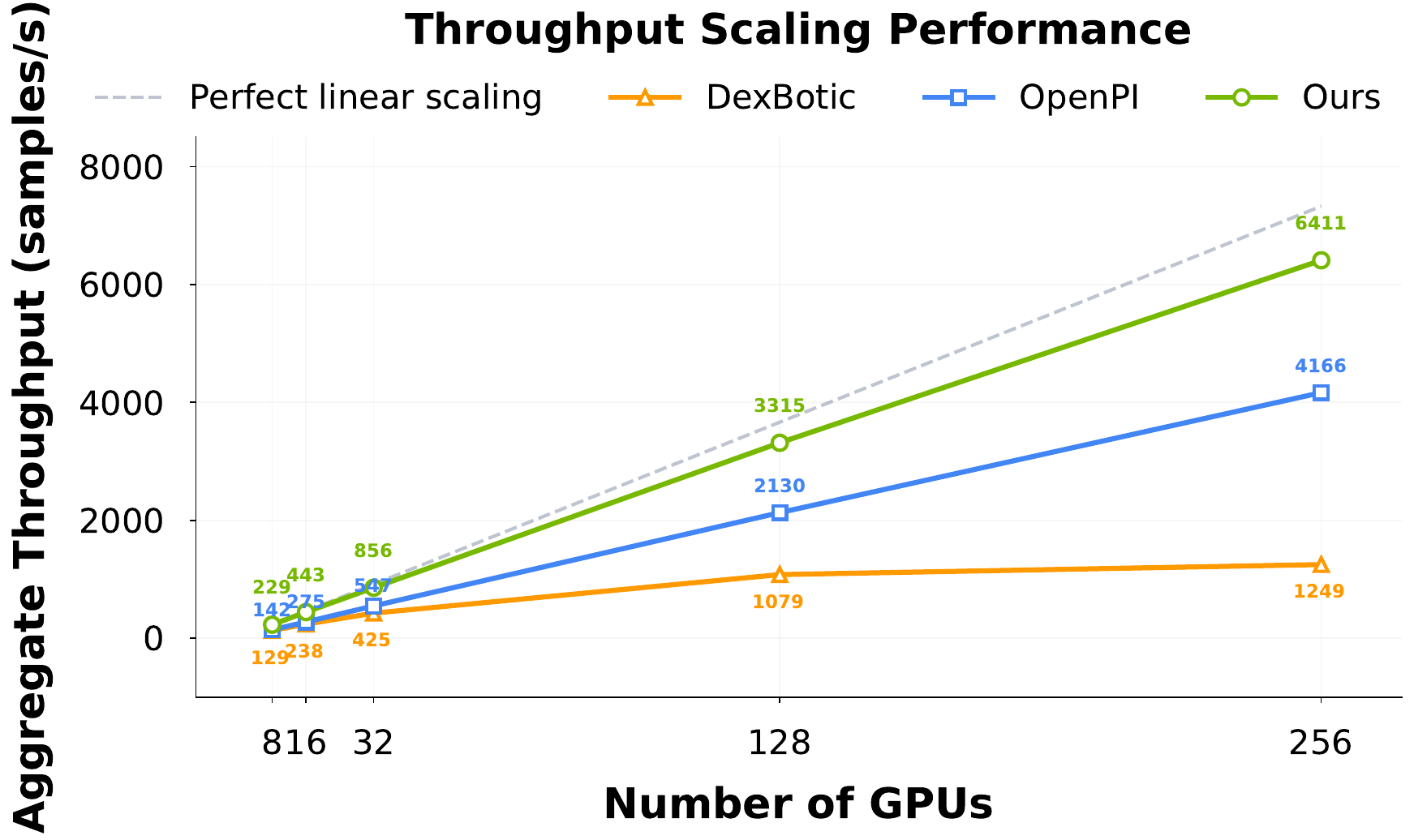}
        \subcaption{PaliGemma-3B-pt-224-$\pi$ model}
        \label{fig2:tp_paligemma}
    \end{minipage}
    \caption{
    \textbf{Training throughput analysis} of the (a) Qwen2.5-VL-3B-$\pi$ and (b) PaliGemma-3B-pt-224-$\pi$ models.}
\end{figure}

\subsection{Training Throughput Analysis}
To comprehensively evaluate the training efficiency of VLA models across different frameworks, we selected three open-source codebases (\textit{i.e.,} StarVLA, Dexbotic, and OpenPI) as the baselines for comparison. 
To ensure a fair comparison, all experiments were conducted on the Libero dataset using a standardized $\pi$-like model architecture.

Given the variations in the VLM implementations across different codebases, we reproduced both Qwen2.5-VL-3B-$\pi$ and PaliGemma-3B-pt-224-$\pi$ models within our own codebase to facilitate a direct alignment and comparison with the baselines. Regarding the training configuration, the local batch size was standardized to 32 for all experiments.

It is worth noting that while StarVLA and Dexbotic default to ZeRO for distributed training, our codebase employs the comparable FSDP2 strategy. In contrast, OpenPI utilizes DDP, which inherently incurs lower communication overhead. We adopted sample throughput (samples/s) as the primary evaluation metric.

\Cref{fig1:tp_qwenvl,fig2:tp_paligemma} illustrate the training efficiency comparison between our codebase and the baselines for the Qwen2.5-VL-3B-$\pi$ and PaliGemma-3B-pt-224-$\pi$ models, respectively. The results demonstrate that our codebase achieved the fastest training speeds in both model settings. Furthermore, the figures detail the training throughput across configurations of 8, 16, 32, 128, and 256 GPUs, alongside the theoretical linear scaling limit. The data indicates that our solution not only delivers superior throughput but also exhibits excellent scaling efficiency that closely follows the theoretical limit as the number of GPUs increases.


\subsection{Ablation Studies}
\subsubsection{Scaling Experiments}
\begin{figure}[t]
    \centering
    \begin{minipage}{0.48\linewidth}
        \centering
        \includegraphics[width=\linewidth]{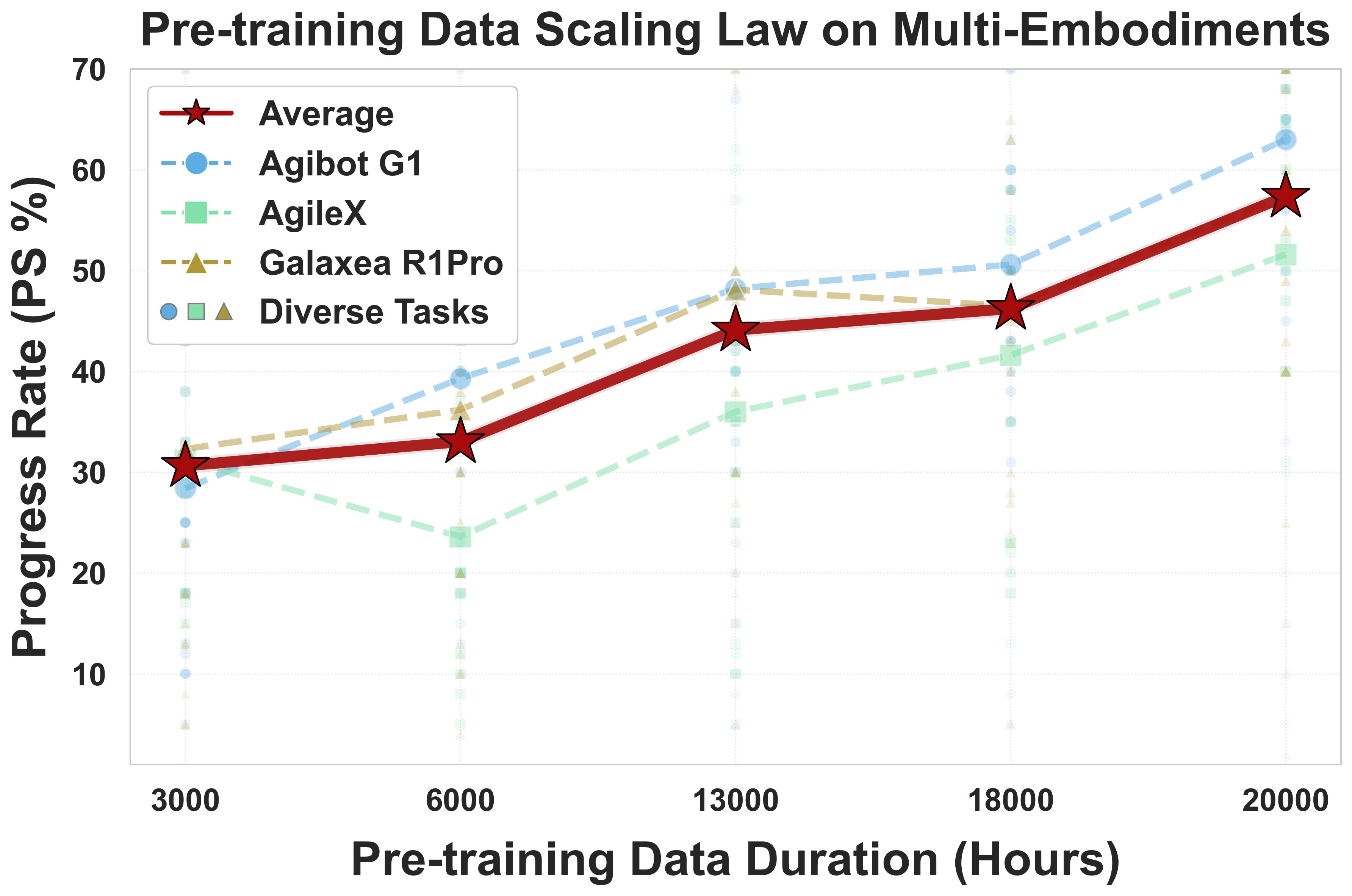}
        \subcaption{Progress Rate (PS)}
        \label{fig1:Progress_Rate_(PS)}
    \end{minipage}
    \hfill
    \begin{minipage}{0.48\linewidth}
        \centering
        \includegraphics[width=\linewidth]{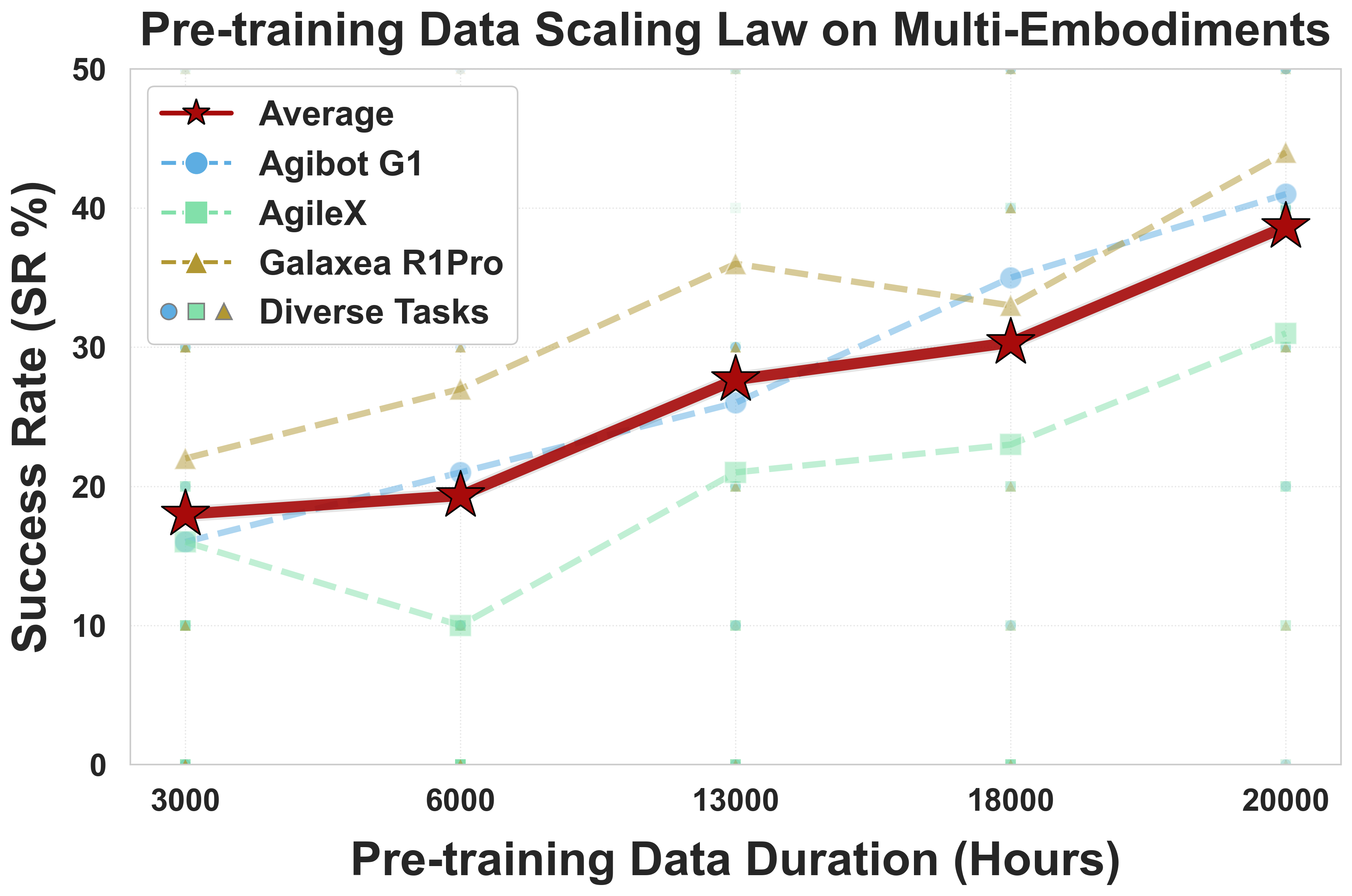}
        \subcaption{Success Rate (SR)}
        \label{fig2:scaling_law_sr}
    \end{minipage}
    \caption{\textbf{Scaling behavior across dataset size.} With increased data scale, our model exhibits scaling laws in terms of success rate and progress rate.}
    \label{fig:scaling_law}
\end{figure}

To assess the scaling laws of pre-training data, we conduct experiments on a subset of 25 representative tasks drawn from the benchmark.
As shown in~\cref{fig1:Progress_Rate_(PS),fig2:scaling_law_sr}, both the progress rate and success rate demonstrate a consistent upward trend as the pre-training data duration increases from 3,000 to 20,000 hours.
This indicates that scaling up real-world pre-training data contributes to improved generalization and performance across diverse downstream tasks and embodiments.
Furthermore, the individual trends of the three embodiments (\textit{i.e.,} Agibot G1, AgileX, and Galaxea R1Pro) generally align with the aggregated performance, suggesting the observed scaling law is robust and not specific to a single platform. These results validate the effectiveness of our scaling approach in enhancing the capabilities of the generalist policy.

\subsubsection{Data-efficient Analysis}

\begin{wrapfigure}{r}{0.48\linewidth} 
    \centering
    \vspace{-31pt}
    \includegraphics[width=0.85\linewidth]{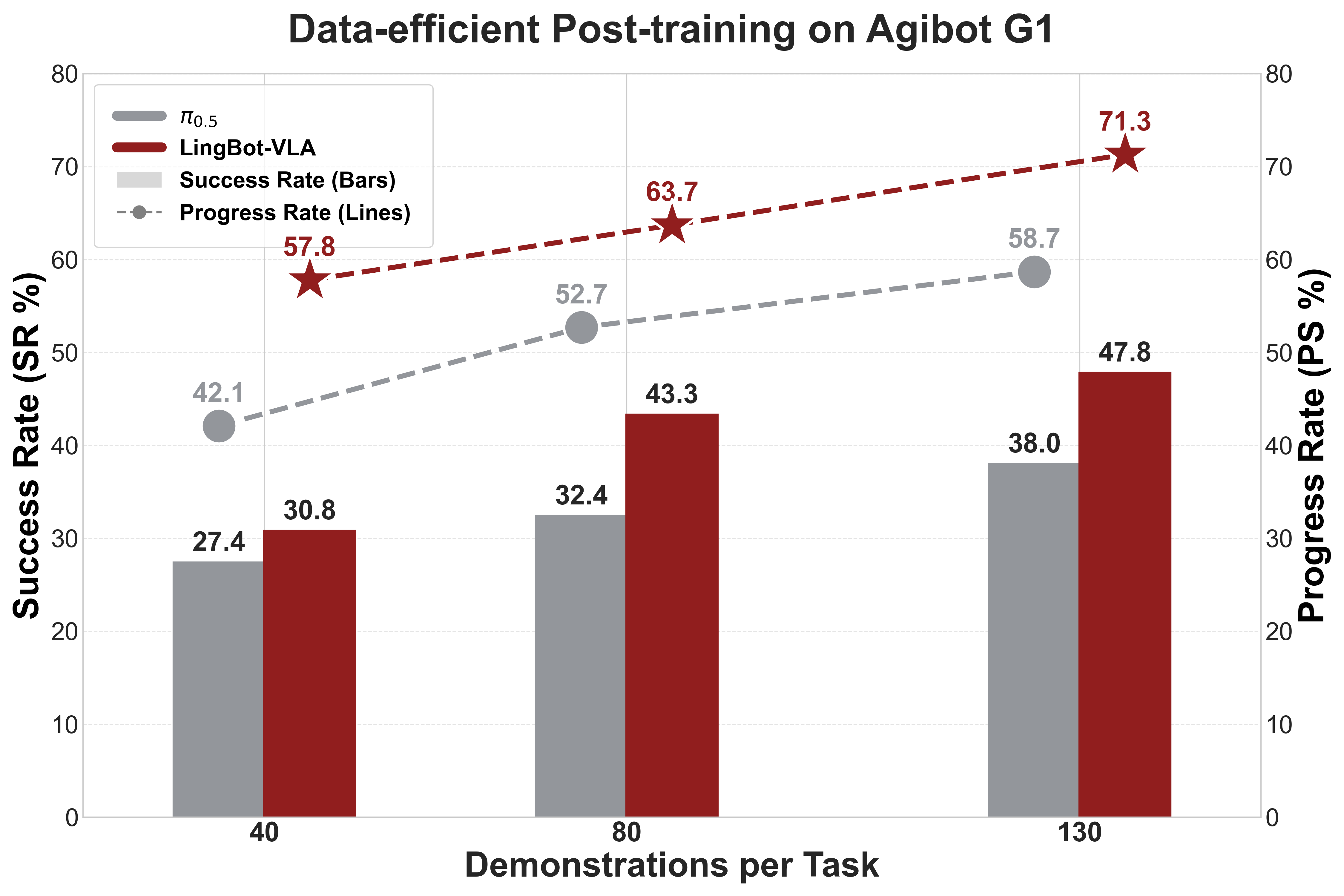}
    \caption{\textbf{Data efficiency} of \method post-training.}
    \label{fig:data_efficient_post_training}
\end{wrapfigure}
Following the large-scale real-world benchmarking protocols, we selected eight representative tasks from GM-100 dataset to conduct data-efficient post-training experiments on the Agibot G1 platform. As illustrated in~\cref{fig:data_efficient_post_training}, with a limited budget of only 80 demonstrations per task, \method outperforms $\pi_{0.5}$ using the full 130-demonstration set in both Progress Rate and Success Rate. Notably, the performance margin between \method and $\pi_{0.5}$ widens significantly as the volume of post-training data increases, demonstrating superior data efficiency and scalability.

\section{Conclusion}\label{sec:conclusion}

We have introduced \method, a foundation model that achieves superior generalizability and training efficiency through large-scale real-world data and an optimized codebase. 
Our comprehensive evaluation across 100 tasks demonstrates that our model achieves clear superiority over competitors, showcasing its strong performance and broad generalizability.
To foster open science, we release our code, model, and benchmark data.
Future research will focus on scaling the model versatility by integrating single-arm and mobile robotic data, paving the way for more diverse and mobile manipulation capabilities in unconstrained environments.

\noindent\textbf{Acknowledgment.}
We thank Zhengyu He, Han Zhang, Yuhao Xie, Haidan Zhou, Chongjun Zhong, Yida Zou, Siyuan Li, Zhikun Luo, Yuanqi Chen, Yingying Zhang, Yijun Zheng, Wanting Xu, Hongfei Niu, Yan Zha, Ka Leong Cheng, Liping Zhang, Zhen Liu, Rundong Zhou, Yuan Guan, Haitao Wang, Weilun Yao, Zhiwei Liang, Jiahao Fan, Jingran Xu, Linyu Su, Haiwei Liang, Yixiang Gao, Yingmin Li, Yongqiang Wen, Jiafei Li, Yuanzhe Guo, Yijun Zheng, Fengrui Zhang, Lin Wang, Min Yao, Fei Lu, Jingyun Tian, Ting Huang, Xinyang Wang, Jianxue Qian, Wenhui Shi, Jie Ren, Lina Tao, Yongqiang Wu, Daohua Yan, Zhou Yang, Qiang Wang and Yao Su for help with data, evaluation experiments, training infrastructure, robot hardware and robot software.
We also gratefully acknowledge Galaxea Team, AgileX Robotics, and Leju(Shenzhen) Robotics Technology Co., Ltd. for help with the data collection and benchmark.

{
\small
\bibliographystyle{plain}
\bibliography{ref.bib}
}

\appendix
\renewcommand\thesection{\Alph{section}}
\renewcommand\thefigure{S\arabic{figure}}
\renewcommand\thetable{S\arabic{table}}
\renewcommand\theequation{S\arabic{equation}}
\setcounter{figure}{0}
\setcounter{table}{0}
\setcounter{equation}{0}

\section*{Appendix}

\section{Experiment}\label{appendix:overview}
This section provides a comprehensive breakdown of the experimental results. 
Specifically, \Cref{tab:exp_AgilexCobotMagic_1}, \Cref{tab:exp_AgilexCobotMagic_2}, \Cref{tab:exp_a2d_1}, \Cref{tab:exp_a2d_2}, \Cref{tab:exp_GalaxeaR1Pro_1} and \Cref{tab:exp_GalaxeaR1Pro_2} below present the detailed performance on GM-100 real-world benchmark. \Cref{tab:robotwin} presents the detailed performance on Robotwin 2.0 benchmark.
These tables serve as the basis for the aggregated mean results reported in the main text (see~\cref{sec:real} and~\cref{sec:sim}).

\begin{table}[t]
  \centering
  \caption{\textbf{Real-world evaluation} on GM-100~\cite{gm100} benchmark (\textbf{Wheeled Humanoid Robot}: AgileX, Part I).}
  \small
  \setlength{\tabcolsep}{12pt}
\begin{tabular}{ccccccccccc}
    \toprule
    \multirow{2}[2]{*}{\textbf{Tasks}} & \multicolumn{2}{c}{WALL-OSS} & \multicolumn{2}{c}{GR00T N1.6} & \multicolumn{2}{c}{$\pi_{0.5}$} & \multicolumn{2}{c}{Ours \textit{w/o} depth} & \multicolumn{2}{c}{Ours \textit{w/} depth} \\
          & \textbf{SR} & \textbf{PS} & \textbf{SR} & \textbf{PS} & \textbf{SR} & \textbf{PS} & \textbf{SR} & \textbf{PS} & \textbf{SR} & \textbf{PS} \\
    \midrule
    \textit{\#001} & 7\%   & 17\%  & 20\%  & 23\%  & 53\%  & 57\%  & 40\%  & 67\%  & 47\%  & 73\% \\
    \textit{\#002} & 0\%   & 3\%   & 0\%   & 0\%   & 0\%   & 23\%  & 0\%   & 15\%  & 0\%   & 2\% \\
    \textit{\#003} & 0\%   & 1\%   & 0\%   & 0\%   & 20\%  & 53\%  & 7\%   & 42\%  & 0\%   & 19\% \\
    \textit{\#006} & 0\%   & 3\%   & 0\%   & 0\%   & 0\%   & 20\%  & 0\%   & 0\%   & 0\%   & 17\% \\
    \textit{\#007} & 0\%   & 16\%  & 0\%   & 7\%   & 0\%   & 41\%  & 20\%  & 61\%  & 27\%  & 70\% \\
    \textit{\#008} & 0\%   & 8\%   & 0\%   & 0\%   & 0\%   & 1\%   & 0\%   & 39\%  & 0\%   & 66\% \\
\bottomrule
    \end{tabular}%
  \label{tab:exp_AgilexCobotMagic_1}%
\end{table}%

\begin{table}[t]
  \centering
  \caption{\textbf{Real-world evaluation} on GM-100~\cite{gm100} benchmark (\textbf{Wheeled Humanoid Robot}: AgileX, Part II).}
  \small
  \setlength{\tabcolsep}{12pt}
    \begin{tabular}{ccccccccccc}
    \toprule
    \multirow{2}[2]{*}{\textbf{Tasks}} & \multicolumn{2}{c}{WALL-OSS} & \multicolumn{2}{c}{GR00T N1.6} & \multicolumn{2}{c}{$\pi_{0.5}$} & \multicolumn{2}{c}{Ours \textit{w/o} depth} & \multicolumn{2}{c}{Ours \textit{w/} depth} \\
          & \textbf{SR} & \textbf{PS} & \textbf{SR} & \textbf{PS} & \textbf{SR} & \textbf{PS} & \textbf{SR} & \textbf{PS} & \textbf{SR} & \textbf{PS} \\
    \midrule
\textit{\#056} & 0\%   & 28\%  & 0\%   & 7\%   & 67\%  & 88\%  & 73\%  & 93\%  & 80\%  & 92\% \\
    \textit{\#057} & 0\%   & 2\%   & 0\%   & 1\%   & 0\%   & 13\%  & 0\%   & 19\%  & 0\%   & 27\% \\
    \textit{\#058} & 0\%   & 37\%  & 7\%   & 44\%  & 7\%   & 58\%  & 7\%   & 63\%  & 20\%  & 72\% \\
    \textit{\#060} & 0\%   & 8\%   & 20\%  & 42\%  & 93\%  & 93\%  & 27\%  & 57\%  & 13\%  & 50\% \\
    \textit{\#061} & 53\%  & 77\%  & 7\%   & 33\%  & 47\%  & 68\%  & 93\%  & 97\%  & 93\%  & 97\% \\
    \textit{\#062} & 0\%   & 2\%   & 0\%   & 0\%   & 0\%   & 8\%   & 0\%   & 4\%   & 0\%   & 1\% \\
\bottomrule
    \end{tabular}%
  \label{tab:exp_AgilexCobotMagic_2}%
\end{table}%

\begin{table}[t]
  \centering
  \caption{\textbf{Real-world evaluation} on GM-100~\cite{gm100} benchmark (\textbf{Wheeled Humanoid Robot}: AgibotG1, Part I).}
  \small
  \setlength{\tabcolsep}{12pt}
    \begin{tabular}{ccccccccccc}
    \toprule
    \multirow{2}[2]{*}{\textbf{Tasks}} & \multicolumn{2}{c}{WALL-OSS} & \multicolumn{2}{c}{GR00T N1.6} & \multicolumn{2}{c}{$\pi_{0.5}$} & \multicolumn{2}{c}{Ours \textit{w/o} depth} & \multicolumn{2}{c}{Ours \textit{w/} depth} \\
          & \textbf{SR} & \textbf{PS} & \textbf{SR} & \textbf{PS} & \textbf{SR} & \textbf{PS} & \textbf{SR} & \textbf{PS} & \textbf{SR} & \textbf{PS} \\
    \midrule
    \textit{\#001} & 7\%   & 20\%  & 0\%   & 0\%   & 0\%   & 23\%  & 0\%   & 3\%   & 0\%   & 17\% \\
    \textit{\#002} & 0\%   & 0\%   & 0\%   & 0\%   & 13\%  & 35\%  & 0\%   & 0\%   & 0\%   & 5\% \\
    \textit{\#003} & 0\%   & 3\%   & 0\%   & 3\%   & 0\%   & 35\%  & 7\%   & 19\%  & 0\%   & 0\% \\
    \textit{\#006} & 0\%   & 0\%   & 0\%   & 0\%   & 0\%   & 0\%   & 0\%   & 0\%   & 0\%   & 0\% \\
    \textit{\#007} & 0\%   & 0\%   & 0\%   & 18\%  & 7\%   & 44\%  & 0\%   & 25\%  & 13\%  & 52\% \\
    \textit{\#008} & 0\%   & 0\%   & 0\%   & 10\%  & 0\%   & 7\%   & 0\%   & 2\%   & 0\%   & 0\% \\
\bottomrule
    \end{tabular}%
  \label{tab:exp_a2d_1}%
\end{table}%

\begin{table}[t]
  \centering
  \caption{\textbf{Real-world evaluation} on GM-100~\cite{gm100} benchmark (\textbf{Wheeled Humanoid Robot}: AgibotG1, Part II).}
  \small
  \setlength{\tabcolsep}{12pt}
    \begin{tabular}{ccccccccccc}
    \toprule
    \multirow{2}[2]{*}{\textbf{Tasks}} & \multicolumn{2}{c}{WALL-OSS} & \multicolumn{2}{c}{GR00T N1.6} & \multicolumn{2}{c}{$\pi_{0.5}$} & \multicolumn{2}{c}{Ours \textit{w/o} depth} & \multicolumn{2}{c}{Ours \textit{w/} depth} \\
          & \textbf{SR} & \textbf{PS} & \textbf{SR} & \textbf{PS} & \textbf{SR} & \textbf{PS} & \textbf{SR} & \textbf{PS} & \textbf{SR} & \textbf{PS} \\
    \midrule
    \textit{\#056} & 0\%   & 0\%   & 0\%   & 17\%  & 0\%   & 15\%  & 7\%   & 29\%  & 7\%   & 22\% \\
    \textit{\#057} & 0\%   & 0\%   & 0\%   & 1\%   & 0\%   & 3\%   & 0\%   & 13\%  & 0\%   & 8\% \\
    \textit{\#058} & 0\%   & 32\%  & 0\%   & 27\%  & 20\%  & 77\%  & 7\%   & 63\%  & 27\%  & 75\% \\
    \textit{\#060} & 7\%   & 38\%  & 0\%   & 22\%  & 27\%  & 67\%  & 0\%   & 37\%  & 0\%   & 17\% \\
    \textit{\#061} & 47\%  & 50\%  & 0\%   & 3\%   & 40\%  & 67\%  & 93\%  & 97\%  & 60\%  & 80\% \\
    \textit{\#062} & 0\%   & 0\%   & 0\%   & 2\%   & 0\%   & 2\%   & 0\%   & 2\%   & 0\%   & 2\% \\
\bottomrule
    \end{tabular}%
  \label{tab:exp_a2d_2}%
\end{table}%

\begin{table}[t]
  \centering
  \caption{\textbf{Real-world evaluation} on GM-100~\cite{gm100} benchmark (\textbf{Wheeled Humanoid Robot}: Galaxea R1Pro, Part I).}
  \small
  \setlength{\tabcolsep}{12pt}
    \begin{tabular}{ccccccccccc}
    \toprule
    \multirow{2}[2]{*}{\textbf{Tasks}} & \multicolumn{2}{c}{WALL-OSS} & \multicolumn{2}{c}{GR00T N1.6} & \multicolumn{2}{c}{$\pi_{0.5}$} & \multicolumn{2}{c}{Ours \textit{w/o} depth} & \multicolumn{2}{c}{Ours \textit{w/} depth} \\
          & \textbf{SR} & \textbf{PS} & \textbf{SR} & \textbf{PS} & \textbf{SR} & \textbf{PS} & \textbf{SR} & \textbf{PS} & \textbf{SR} & \textbf{PS} \\
    \midrule
    \textit{\#001} & 13\%  & 17\%  & 13\%  & 20\%  & 20\%  & 43\%  & 20\%  & 27\%  & 27\%  & 47\% \\
    \textit{\#002} & 33\%  & 45\%  & 0\%   & 2\%   & 0\%   & 2\%   & 0\%   & 10\%  & 0\%   & 8\% \\
    \textit{\#003} & 7\%   & 49\%  & 0\%   & 0\%   & 0\%   & 0\%   & 73\%  & 89\%  & 33\%  & 44\% \\
    \textit{\#006} & 0\%   & 1\%   & 0\%   & 12\%  & 0\%   & 0\%   & 7\%   & 28\%  & 0\%   & 6\% \\
    \textit{\#007} & 0\%   & 2\%   & 0\%   & 8\%   & 13\%  & 18\%  & 0\%   & 0\%   & 33\%  & 39\% \\
    \textit{\#008} & 0\%   & 4\%   & 0\%   & 14\%  & 0\%   & 20\%  & 0\%   & 43\%  & 0\%   & 45\% \\
\bottomrule
    \end{tabular}%
  \label{tab:exp_GalaxeaR1Pro_1}%
\end{table}%

\begin{table}[t]
  \centering
  \caption{\textbf{Real-world evaluation} on GM-100~\cite{gm100} benchmark (\textbf{Wheeled Humanoid Robot}: Galaxea R1Pro, Part II).}
  \small
  \setlength{\tabcolsep}{12pt}
    \begin{tabular}{ccccccccccc}
    \toprule
    \multirow{2}[2]{*}{\textbf{Tasks}} & \multicolumn{2}{c}{WALL-OSS} & \multicolumn{2}{c}{GR00T N1.6} & \multicolumn{2}{c}{$\pi_{0.5}$} & \multicolumn{2}{c}{Ours \textit{w/o} depth} & \multicolumn{2}{c}{Ours \textit{w/} depth} \\
          & \textbf{SR} & \textbf{PS} & \textbf{SR} & \textbf{PS} & \textbf{SR} & \textbf{PS} & \textbf{SR} & \textbf{PS} & \textbf{SR} & \textbf{PS} \\
    \midrule
\textit{\#056} & 0\%   & 8\%   & 7\%   & 32\%  & 0\%   & 22\%  & 0\%   & 52\%  & 47\%  & 68\% \\
    \textit{\#057} & 0\%   & 0\%   & 0\%   & 4\%   & 0\%   & 0\%   & 0\%   & 38\%  & 0\%   & 3\% \\
    \textit{\#058} & 0\%   & 8\%   & 0\%   & 15\%  & 7\%   & 15\%  & 20\%  & 30\%  & 20\%  & 33\% \\
    \textit{\#060} & 0\%   & 15\%  & 7\%   & 47\%  & 7\%   & 53\%  & 20\%  & 58\%  & 20\%  & 68\% \\
    \textit{\#061} & 47\%  & 57\%  & 60\%  & 63\%  & 47\%  & 57\%  & 87\%  & 93\%  & 93\%  & 97\% \\
    \textit{\#062} & 0\%   & 5\%   & 0\%   & 2\%   & 0\%   & 3\%   & 0\%   & 0\%   & 0\%   & 1\% \\
\bottomrule
    \end{tabular}%
  \label{tab:exp_GalaxeaR1Pro_2}%
\end{table}%

\begin{table}[t]
  \centering
  \caption{\textbf{Real-world evaluation} on GM-100~\cite{gm100} benchmark (\textbf{Bipedal Humanoid Robot}: Leju KUAVO 4 Pro, Part I).}
  \small
  \setlength{\tabcolsep}{12pt}
    \begin{tabular}{ccccccccccc}
    \toprule
      \multirow{2}[2]{*}{\textbf{Tasks}} & \multicolumn{2}{c}{WALL-OSS} & \multicolumn{2}{c}{GR00T N1.6} &
  \multicolumn{2}{c}{$\pi_{0.5}$} & \multicolumn{2}{c}{Ours \textit{w/o} depth} & \multicolumn{2}{c}{Ours
  \textit{w/} depth} \\
            & \textbf{SR} & \textbf{PS} & \textbf{SR} & \textbf{PS} & \textbf{SR} & \textbf{PS} & \textbf{SR} &
  \textbf{PS} & \textbf{SR} & \textbf{PS} \\
    \midrule
      \textit{\#001} & 0\% & 3\% & 0\% & 13\% & 67\% & 68\% & 80\% & 85\% & 80\% & 82\% \\
      \textit{\#002} & 0\% & 0\% & 0\% & 2\% & 0\% & 25\% & 0\% & 22\% & 0\% & 27\% \\
      \textit{\#003} & 0\% & 3\% & 7\% & 27\% & 73\% & 80\% & 80\% & 87\% & 73\% & 83\% \\
      \textit{\#004} & 0\% & 2\% & 0\% & 0\% & 0\% & 0\% & 0\% & 11\% & 0\% & 16\% \\
      \textit{\#005} & 0\% & 4\% & 0\% & 8\% & 0\% & 0\% & 0\% & 20\% & 0\% & 11\% \\
      \textit{\#006} & 0\% & 3\% & 0\% & 2\% & 0\% & 8\% & 7\% & 15\% & 0\% & 18\% \\
\bottomrule
    \end{tabular}%
  \label{tab:exp_LejuKUAVO_1}%
\end{table}%

\begin{table}[t]
  \centering
  \caption{\textbf{Real-world evaluation} on GM-100~\cite{gm100} benchmark (\textbf{Bipedal Humanoid Robot}: Leju KUAVO 4 Pro, Part II).}
  \small
  \setlength{\tabcolsep}{12pt}
    \begin{tabular}{ccccccccccc}
    \toprule
      \multirow{2}[2]{*}{\textbf{Tasks}} & \multicolumn{2}{c}{WALL-OSS} & \multicolumn{2}{c}{GR00T N1.6} &
    \multicolumn{2}{c}{$\pi_{0.5}$} & \multicolumn{2}{c}{Ours \textit{w/o} depth} & \multicolumn{2}{c}{Ours
    \textit{w/} depth} \\
  & \textbf{SR} & \textbf{PS} & \textbf{SR} & \textbf{PS} & \textbf{SR} & \textbf{PS} & \textbf{SR} &
  \textbf{PS} & \textbf{SR} & \textbf{PS} \\
    \midrule
      \textit{\#059} & 0\% & 67\% & 0\% & 48\% & 7\% & 70\% & 33\% & 80\% & 13\% & 70\% \\
      \textit{\#060} & 73\% & 87\% & 87\% & 93\% & 100\% & 100\% & 87\% & 93\% & 93\% & 97\% \\
      \textit{\#061} & 0\% & 0\% & 0\% & 1\% & 0\% & 0\% & 0\% & 0\% & 0\% & 0\% \\
      \textit{\#062} & 0\% & 13\% & 13\% & 33\% & 13\% & 53\% & 0\% & 30\% & 0\% & 10\% \\
      \textit{\#063} & 0\% & 0\% & 0\% & 0\% & 0\% & 0\% & 0\% & 0\% & 0\% & 2\% \\
      \textit{\#064} & 0\% & 0\% & 0\% & 15\% & 0\% & 27\% & 0\% & 0\% & 0\% & 15\% \\
\bottomrule
    \end{tabular}%
  \label{tab:exp_LejuKUAVO_2}%
\end{table}%

\begin{table}[t]
  \centering
  \caption{\textbf{Simulation evaluation} on RoboTwin 2.0~\cite{robotwin2} benchmark, under ``clean'' and ``randomized'' settings.}
  \small
  \setlength{\tabcolsep}{15.5pt}
    \begin{tabular}{ccccccc}
    \toprule
    \multicolumn{1}{c}{\multirow{2}[2]{*}{\textbf{Simulation Tasks}}} & \multicolumn{2}{c}{\textbf{$\pi_{0.5}$}} & \multicolumn{2}{c}{Ours \textit{w/o} depth} & \multicolumn{2}{c}{Ours \textit{w/} depth} \\
          & \textbf{Clean} & \textbf{Rand.} & \textbf{Clean} & \textbf{Rand.} & \textbf{Clean} & \textbf{Rand.} \\
    \midrule
    \textit{Adjust Bottle} & 100\%   & 99\%    & 100\%   & 100\%   & 100\%   & 100\% \\
    \textit{Beat Block Hammer} & 96\%    & 93\%    & 87\%    & 91\%    & 92\%    & 89\% \\
    \textit{Blocks Ranking Rgb} & 92\%    & 85\%    & 92\%    & 91\%    & 92\%    & 91\% \\
    \textit{Blocks Ranking Size} & 49\%    & 26\%    & 66\%    & 73\%    & 76\%    & 70\% \\
    \textit{Click Alarmclock} & 98\%    & 89\%    & 93\%    & 26\%    & 97\%    & 43\% \\
    \textit{Click Bell} & 99\%    & 66\%    & 32\%    & 19\%    & 43\%    & 36\% \\
\bottomrule
    \end{tabular}
  \label{tab:robotwin}%
\end{table}%

\end{document}